\documentclass{article} % For LaTeX2e
\usepackage{iclr2015,times}
\usepackage{hyperref}
\usepackage{url}
\usepackage{cleveref}
\usepackage{amsmath}
\usepackage{amssymb}
\usepackage{tikz}
\usepackage{booktabs}
\usepackage{caption}
\usepackage{subcaption}

\newcommand{\permuscale}{.6}

\usetikzlibrary{calc,arrows,backgrounds}

\definecolor{ph-blue}{rgb}{0, 0, 0.509}
\definecolor{ph-gray}{rgb}{0.5, 0.5, 0.5}
\definecolor{ph-oran}{rgb}{1, 0.6, 0.2}
\definecolor{ph-dred}{rgb}{0.4906, 0, 0}

\definecolor{ph-light-gray}{rgb}{0.75, 0.75, 0.75}

\title{Permutohedral Lattice CNNs}

\author{
Martin Kiefel, Varun Jampani and Peter V. Gehler\\
Max Planck Institute for Intelligent Systems\\
T{\"u}bingen, 72076, Germany\\
\texttt{\{martin.kiefel, varun.jampani, peter.gehler\}@tuebingen.mpg.de} \\
}

\iclrfinalcopy % Uncomment for camera-ready version

\begin{document}

\maketitle

\begin{abstract}
  This paper presents a convolutional layer that is able to process
  sparse input features. As an example, for image recognition problems
  this allows an efficient filtering of signals that do not lie on a
  dense grid (like pixel position), but of more general features (such
  as color values). The presented algorithm makes use of the
  permutohedral lattice data structure. The permutohedral lattice was
  introduced to efficiently implement a bilateral filter, a commonly
  used image processing operation. Its use allows for a generalization
  of the convolution type found in current (spatial) convolutional
  network architectures.
\end{abstract}

\section{Introduction}
In the domain of image recognition, the convolutional layer of a CNN
today is almost exclusively associated with a spatial convolution in
the image domain. In this work we will take a more signal theoretic
viewpoint of the convolutional operation and present an algorithm that
allows to process also sparse input data. This work is inspired by the
use of special data structures~\citep{adams10eurographics} for bilateral
filters~\citep{aurich95dagm, smith97ijcv, tomasi98iccv} and
generalizes it for the use of convolutional architectures.

Although the approach presented here is more general, the following
two scenarios are instructive. Consider that at training time we have
access to full resolution images to train a classifier. At test time
only a random number of pixels from the test image is available. In
other words, we sample the signal differently during training and test
time. For a traditional CNN this would require a pre-processing step,
for example to map from subsets of pixels to a dense grid that is the
image. In our view there is no change, it is not required that we have
a dense grid and access to all pixels of the image. That is the
integration domain does not change. This is one example of sparsity,
here we deal with a set of pixels, whose values are RGB and features
are position. Similarly, color information can be used to define the
filtering operation as well. One can devise a convolution with a
domain respecting color \emph{and} location information (or color
alone). One view from the image processing community is that of an
\emph{edge-aware} filter, the filter will be adaptive to the color
and/or gradient of the image. RGB values do not lie on a regular dense
grid, therefore a direct expansion of the spatial convolution is not
applicable.

% Extends the idea of convolution to express the learning of a data
% invariant that we want to model.

% We make use of the Permutohedral lattice~\citep{adams}. All data is projected
% onto a lattice plane to do the blurring operation.

% This has nice applications: We are not restricted any more to get the
% image sampled from a grid. Heck, it can be randomly sampled (point to
% experiments). Show that there are actual applications where this is
% the case: MRT frequency space sampling might be one. Are there others?

This approach falls into line with the view on encoding
invariants~\citep{mallat12pam}. It is possible to encode our knowledge
invariants that we have about the data with the new way of
looking at the data. Encoded in a spatial convolution is the prior
knowledge about translation invariance. How to encode roation
invariance, how similarity in color space? In the view we take here
these are simply convolutions over different domains. A grid based
convolution cannot easily be used to work with the sparse data (an
interpolation might be needed) but the permutohedral lattice provides
the right space and allows efficient implementations.
%% This lattice does not have the problem of the ``curse of dimensionality'' and allows efficient implementations.
Therefore the runtime is comparable to the ones of spatial convolutions,
depending on the size of the invariants to include and can simply be
used as a replacement of the traditional layers.

% Talk about related work. So far Permutohedral lattice has been applied
% to Gaussian blur in graphics (as bilateral filter), and for inference
% in machine learning (as method to do a mean field approximation).
% Interestingly, method is not restricted to Gaussian blurring.

\section{Permutohedral Lattice Convolution}
\begin{figure}[t]
  \colorlet{grid}{ph-gray}
  \colorlet{lightgrid}{ph-light-gray}
  \colorlet{points}{ph-oran}
  \colorlet{arrowscolor}{ph-dred}
  \colorlet{filtercolor}{black}
  \colorlet{filtergridcolor}{ph-dred}
  \centering
  \begin{tikzpicture}[scale=\permuscale, >=stealth']
    %%%  define vertices with coordinates
    \coordinate (0;0) at (0,0);
    \foreach \c in {1,...,3}{%
      \foreach \i in {0,...,5}{%
        \pgfmathtruncatemacro\j{\c*\i}
        \coordinate (\c;\j) at (60*\i:\c);
    } }
    \foreach \i in {0,2,...,10}{%
      \pgfmathtruncatemacro\j{mod(\i+2,12)}
      \pgfmathtruncatemacro\k{\i+1}
      \coordinate (2;\k) at ($(2;\i)!.5!(2;\j)$) ;}

    \foreach \i in {0,3,...,15}{%
      \pgfmathtruncatemacro\j{mod(\i+3,18)}
      \pgfmathtruncatemacro\k{\i+1}
      \pgfmathtruncatemacro\l{\i+2}
      \coordinate (3;\k) at ($(3;\i)!1/3!(3;\j)$)  ;
      \coordinate (3;\l) at ($(3;\i)!2/3!(3;\j)$)  ;
    }

    %%%%%%%%% draw lines %%%%%%%%
    \foreach \i in {0,...,6}{%
      \pgfmathtruncatemacro\k{\i}
      \pgfmathtruncatemacro\l{15-\i}
      \draw[thin,grid] (3;\k)--(3;\l);
      \pgfmathtruncatemacro\k{9-\i}
      \pgfmathtruncatemacro\l{mod(12+\i,18)}
      \draw[thin,grid] (3;\k)--(3;\l);
      \pgfmathtruncatemacro\k{12-\i}
      \pgfmathtruncatemacro\l{mod(15+\i,18)}
      \draw[thin,grid] (3;\k)--(3;\l);}
    %%%%%%%%% some specific lines %%%%%%%%%%
    % \foreach \i in {0,2,...,10} {
    %   \pgfmathtruncatemacro\j{mod(\i+2,12)}
    %   \draw[thick,dashed] (2;\i)--(2;\j);}
    %%%%%%%%% draw points %%%%%%%%
    \fill [grid] (0;0) circle (2pt);
    \foreach \c in {1,...,3}{%
      \pgfmathtruncatemacro\k{\c*6-1}
      \foreach \i in {0,...,\k}{%
        \fill [lightgrid] (\c;\i) circle (2pt);}}
    \fill [black] (0;0) circle (2pt);
    \fill [black] (1;3) circle (2pt);
    \fill [black] (1;4) circle (2pt);
    %%%%%%%%% some specific points %%%%%%%%%%
    \fill [points] (0.52,-0.35) circle (2pt);
    \fill [points] (1.5,-1.7) circle (2pt);
    \fill [points] (0.75,1.1) circle (2pt);
    \fill [points] (-1.0,1.3) circle (2pt);
    \fill [points] (0.5,-2.3) circle (2pt);
    \fill [points] (-1.5,-1) circle (2pt);
    \fill [points] (-0.5,-0.3) circle (2pt);
    \fill [points] (-1.0,-2) circle (2pt);
    \fill [points] (0.7,-0.3) circle (2pt);
    \fill [points] (-0.3,0.7) circle (2pt);
    \fill [points] (-1,0.3) circle (2pt);
    \fill [points] (1.0,1.4) circle (2pt);
    \fill [points] (-0.7,-1.3) circle (2pt);
    \fill [points] (1.3,-0.7) circle (2pt);
    % \foreach \n in {0,3,...,15}{%
    %   \draw (3;\n) circle (4pt);}
    % \foreach \n in {1,3,...,11}{%
    %   \draw (2;\n) circle (4pt);}
    %%%%%%%%%% arrows %%%%%%%%%%%%
    \draw[->,arrowscolor,shorten >=2pt,shorten <=2pt](-0.5,-0.3)--(0;0);
    \draw[->,arrowscolor,shorten >=2pt,shorten <=2pt](-0.5,-0.3)--(1;3);
    \draw[->,arrowscolor,shorten >=2pt,shorten <=2pt](-0.5,-0.3)--(1;4);

    \node [anchor=north west] at (3;9 |- 3;6) {\Large Splat};
  \end{tikzpicture}
  %
  %%% Convolution.
  \begin{tikzpicture}[scale=\permuscale]
    \begin{scope}
      %%%  define vertices with coordinates
      \coordinate (0;0) at (0,0);
      \foreach \c in {1,...,3}{%
        \foreach \i in {0,...,5}{%
          \pgfmathtruncatemacro\j{\c*\i}
          \coordinate (\c;\j) at (60*\i:\c);
      } }
      \foreach \i in {0,2,...,10}{%
        \pgfmathtruncatemacro\j{mod(\i+2,12)}
        \pgfmathtruncatemacro\k{\i+1}
        \coordinate (2;\k) at ($(2;\i)!.5!(2;\j)$) ;}

      \foreach \i in {0,3,...,15}{%
        \pgfmathtruncatemacro\j{mod(\i+3,18)}
        \pgfmathtruncatemacro\k{\i+1}
        \pgfmathtruncatemacro\l{\i+2}
        \coordinate (3;\k) at ($(3;\i)!1/3!(3;\j)$)  ;
        \coordinate (3;\l) at ($(3;\i)!2/3!(3;\j)$)  ;
      }

      %%%%%%%%% draw lines %%%%%%%%
      \foreach \i in {0,...,6}{%
        \pgfmathtruncatemacro\k{\i}
        \pgfmathtruncatemacro\l{15-\i}
        \draw[thin,grid] (3;\k)--(3;\l);
        \pgfmathtruncatemacro\k{9-\i}
        \pgfmathtruncatemacro\l{mod(12+\i,18)}
        \draw[thin,grid] (3;\k)--(3;\l);
        \pgfmathtruncatemacro\k{12-\i}
        \pgfmathtruncatemacro\l{mod(15+\i,18)}
        \draw[thin,grid] (3;\k)--(3;\l);}
      %%%%%%%%% draw points %%%%%%%%
      \fill [black] (0;0) circle (2pt);
      \foreach \c in {1,...,3}{%
        \pgfmathtruncatemacro\k{\c*6-1}
        \foreach \i in {0,...,\k}{%
          \fill [lightgrid] (\c;\i) circle (2pt);}}

      \node [anchor=north west] at (3;9 |- 3;6) {\Large Convolve};
    \end{scope}

    \begin{scope}[xshift=0cm]
      %%%  define vertices with coordinates
      \coordinate (0;0) at (0,0);
      \foreach \c in {1,...,3}{%
        \foreach \i in {0,...,5}{%
          \pgfmathtruncatemacro\j{\c*\i}
          \coordinate (\c;\j) at (60*\i:\c);
      } }
      \foreach \i in {0,2,...,10}{%
        \pgfmathtruncatemacro\j{mod(\i+2,12)}
        \pgfmathtruncatemacro\k{\i+1}
        \coordinate (2;\k) at ($(2;\i)!.5!(2;\j)$) ;}

      \foreach \i in {0,3,...,15}{%
        \pgfmathtruncatemacro\j{mod(\i+3,18)}
        \pgfmathtruncatemacro\k{\i+1}
        \pgfmathtruncatemacro\l{\i+2}
        \coordinate (3;\k) at ($(3;\i)!1/3!(3;\j)$)  ;
        \coordinate (3;\l) at ($(3;\i)!2/3!(3;\j)$)  ;
      }

      %%%%%%%%% draw lines %%%%%%%%
      \foreach \i in {0,...,4}{%
        \pgfmathtruncatemacro\k{\i}
        \pgfmathtruncatemacro\l{10-\i}
        \draw[thin,filtergridcolor] (2;\k)--(2;\l);
        \pgfmathtruncatemacro\k{6-\i}
        \pgfmathtruncatemacro\l{mod(8+\i,12)}
        \draw[thin,filtergridcolor] (2;\k)--(2;\l);
        \pgfmathtruncatemacro\k{8-\i}
        \pgfmathtruncatemacro\l{mod(10+\i,12)}
        \draw[thin,filtergridcolor] (2;\k)--(2;\l);
      }
      %%%%%%%%% draw points %%%%%%%%
      \fill [filtercolor] (0;0) circle (2pt);
      \foreach \c in {1,...,2}{%
        \pgfmathtruncatemacro\k{\c*6-1}
        \foreach \i in {0,...,\k}{%
          \fill [filtercolor] (\c;\i) circle (2pt);}}
    \end{scope}
  \end{tikzpicture}
  %
  %%% Slice.
  \begin{tikzpicture}[scale=\permuscale, >=stealth']
    %%%  define vertices with coordinates
    \coordinate (0;0) at (0,0);
    \foreach \c in {1,...,3}{%
      \foreach \i in {0,...,5}{%
        \pgfmathtruncatemacro\j{\c*\i}
        \coordinate (\c;\j) at (60*\i:\c);
    } }
    \foreach \i in {0,2,...,10}{%
      \pgfmathtruncatemacro\j{mod(\i+2,12)}
      \pgfmathtruncatemacro\k{\i+1}
      \coordinate (2;\k) at ($(2;\i)!.5!(2;\j)$) ;}

    \foreach \i in {0,3,...,15}{%
      \pgfmathtruncatemacro\j{mod(\i+3,18)}
      \pgfmathtruncatemacro\k{\i+1}
      \pgfmathtruncatemacro\l{\i+2}
      \coordinate (3;\k) at ($(3;\i)!1/3!(3;\j)$)  ;
      \coordinate (3;\l) at ($(3;\i)!2/3!(3;\j)$)  ;
    }

    %%%%%%%%% draw lines %%%%%%%%
    \foreach \i in {0,...,6}{%
      \pgfmathtruncatemacro\k{\i}
      \pgfmathtruncatemacro\l{15-\i}
      \draw[thin,grid] (3;\k)--(3;\l);
      \pgfmathtruncatemacro\k{9-\i}
      \pgfmathtruncatemacro\l{mod(12+\i,18)}
      \draw[thin,grid] (3;\k)--(3;\l);
      \pgfmathtruncatemacro\k{12-\i}
      \pgfmathtruncatemacro\l{mod(15+\i,18)}
      \draw[thin,grid] (3;\k)--(3;\l);}
    %%%%%%%%% draw points %%%%%%%%
    \fill [black] (0;0) circle (2pt);
    \foreach \c in {1,...,3}{%
      \pgfmathtruncatemacro\k{\c*6-1}
      \foreach \i in {0,...,\k}{%
        \fill [lightgrid] (\c;\i) circle (2pt);}}
    \fill [black] (0;0) circle (2pt);
    \fill [black] (1;3) circle (2pt);
    \fill [black] (1;4) circle (2pt);
    %%%%%%%%% some specific points %%%%%%%%%%
    \fill [points] (0.52,-0.35) circle (2pt);
    \fill [points] (1.5,-1.7) circle (2pt);
    \fill [points] (0.75,1.1) circle (2pt);
    \fill [points] (-1.0,1.3) circle (2pt);
    \fill [points] (0.5,-2.3) circle (2pt);
    \fill [points] (-1.5,-1) circle (2pt);
    \fill [points] (-0.5,-0.3) circle (2pt);
    \fill [points] (-1.0,-2) circle (2pt);
    \fill [points] (0.7,-0.3) circle (2pt);
    \fill [points] (-0.3,0.7) circle (2pt);
    \fill [points] (-1,0.3) circle (2pt);
    \fill [points] (1.0,1.4) circle (2pt);
    \fill [points] (-0.7,-1.3) circle (2pt);
    \fill [points] (1.3,-0.7) circle (2pt);
    %%%%%%%%%% arrows %%%%%%%%%%%%
    \draw[->,arrowscolor,shorten >=2pt,shorten <=3pt](0;0)--(-0.5,-0.3);
    \draw[->,arrowscolor,shorten >=2pt,shorten <=3pt](1;3)--(-0.5,-0.3);
    \draw[->,arrowscolor,shorten >=2pt,shorten <=3pt](1;4)--(-0.5,-0.3);

    \node [anchor=north west] at (3;9 |- 3;6) {\Large Slice};
  \end{tikzpicture}

  \caption{The permutohedral convolution consists of three steps:
    first the samples are splatted onto the lattice, then a
    convolution operates on the lattice considering a margin of $s =
    2$ neighbors of a node, and finally the result of the convolution
    is transformed back to the output samples.}
  \label{fig:ops}
  \vspace{-.3cm}
\end{figure}
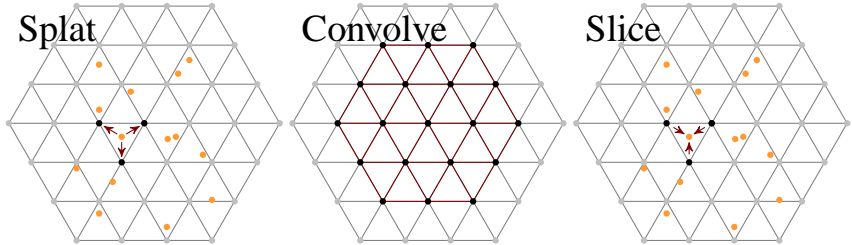

We propose a convolution operation of a $d$-dimensional input space
that entirely works on a lattice. Input data is a tuple $(f_i,v_i)$ of
feature locations $f_i\in\mathbb{R}^d$ and corresponding signal values
$v_i\in\mathbb{R}$. Importantly, this does not assume the feature
locations $f_i$ to be sampled on a regular grid, for example $f_i$ can
be position and RGB value. We then map the input signal to a regular
structure, the so-called permutohedral lattice. A convolution then
operates on the constructed lattice and the result is mapped back to
the output space. Hence, the entire operation consists of three
stages (see \cref{fig:ops}): \emph{splat} (the mapping to the lattice
space), \emph{convolution} and \emph{slice} (the mapping back from the
lattice). This strategy has already been used to implement fast
Gaussian filtering~\citep{paris09ijcv, adams10eurographics,
  adams09siggraph}. Here we generalize it to arbitrary convolutions.

The permutohedral lattice is the result of the projection of the set
$\mathbb{Z}^{d+1}$ onto a plane defined by its orthogonal vector
$\textbf{1}\in\mathbb{R}^{d+1}$. This $d$ dimensional plane is
embedded into $\mathbb{R}^{d+1}$. The lattice points tessellate the
subspace with regular cells. Given a point from the embedding space,
it is efficient to find the enclosing simplex of the projection onto the
plane. We will represent a sparse set of points from $\mathbb{R}^d$ by
a sparse set of simplex corners in the lattice. Importantly, the
number of corners does not grow exponentially with the dimension $d$
as it would for an axis-align simplex representation. We continue to
describe the different parts of the permutohedral convolution.

%% Why does it make sense to use the lattice? Corners of each vertex can
%% be found fast (linear in $d$). This makes splatting and slicing fast.
%% Every triangle has $d+1$ corners. Hence blurring is fast. Even in
%% higher dimensions.

%% Splat and slice can be efficiently implemented with lattice operations
%% and the combination of the two operations correspond to a blur with a
%% variance $d (d+1)^2 / 6$~\citep{adams}.

The splat and slice operations take the role of an interpolation
between the different signal representations. All input samples that
belong to a cell adjacent to lattice point $j$ are summed up and
weighted with the barycentric coordinates to calculate the value
$l_j = \sum_{i\in C(j)} b_{i, j} v_i$. This is the splatting
operation. The barycentric coordinates $b_{i, j}$ depend on both the
corner point $j$ and the feature location $f_i$. The reverse operation
slice finds an output value $v'_k$ by using its barycentric
coordinates inside the lattice simplex and sums over the corner points
$v'_k = \sum_{j\in C(k)} b_{k, j} l'_j$.

%% and combined correspond to a blur with a variance of
%% $d (d+1)^2 / 6$~\citep{adams10eurographics}. This already encodes a
%% spatial invariance as we will see later in \cref{sec:invariant}.

The convolution is then performed on the permutohedral lattice. It
uses a convolution kernel $w_n$ to compute
$l'_{j'} = \sum_{(n, j)\in N(j')} w_n l_{j}$. The convolution kernel
$w_n$ is problem specific and its domain is restricted to the set of
neighboring lattice points $N(j)$. For bilateral filters, this is set
to be a Gaussian filter, here we learn the kernel values using
back-propagation.

The size of the neighborhood takes a similar role as the filter size
(spatial extent) of the grid-based CNN. A transitional convolutional
kernel which considers $s$ sampled points to either side has $(2s
+ 1)^d\in\mathcal{O}(s^d)$ parameters. A comparable filter on the
permutohedral lattice with a $s$ neighborhood has $(s+1)^{d+1} -
s^{d+1} \in\mathcal{O}(s^d)$ elements.

\section{Sparse CNNs and Encoding Invariants}
\label{sec:invariant}
The permutohedral convolution can be used as a new building block in a
CNN architecture. We will omit the derivation of the gradients for the
filter elements with respect to the output of such a new layer due to
space constraints. We will discuss two possible application scenarios.

First, as mentioned before we are free to change the sampling of the
input signal of a lattice-based convolution. The choice of the
sampling is problem specific. Missing measurements or domain specific
sampling techniques that gather more information in highly
discriminant areas are only two possible scenarios. Furthermore, as we
will show in our experiments the method is robust in cases where
train-time sampling and test-time sampling do not coincide.

%% Also, the sampling grid used as an input to the permutohedral
%% convolution layer is not required to have the form as the output
%% sampling. A combination of permutohedral layers and standard discrete
%% convolutions is hence possible. By using a specific sampling as input
%% sampling and an equidistant sampling facing the traditional
%% convolution operator.

Second, the proposed method provides a tool to encode additional data
invariances in a principled way. A common technique to include domain
knowledge is to artificially augment the training set with
deformations that leave the output signal invariant, such as
translations, rotations, or nosiy versions.

A feature mapping $\Phi$ is invariant with respect to a transformation
$L$ and a signal $v$ if $\Phi(v) \approx \Phi(v_L)$. In the case where
$L$ belongs to a set of translations a possible invariant feature is
the convolution with a window function $w$ (given its support has the right size)
$\Phi(v, s) = \int w(t) v(s - t) \text{d} t.$ The same idea can be
applied to the more general case and again calculating a mean with the
help of a window function: $\Phi(v, L) = \int w(M) v(M^{-1}L) \text{d}M$.

%% The bigger the set $\cal{L}$ the more the feature is invariant towards
%% the operation. There is a trade-off here: If this integral ``means''
%% over a big portion of the signal the feature will not be
%% informative. If the region is too small then the feature will not be
%% invariant enough. (Mallat's work builds a pyramid in this case to
%% cover all levels of this trade-off.)

We can use the permutohedral convolution to encode invariances like
rotation and translation. Approximating the above integral by a finite
sum and using lattice points as integration samples we arrive at
$\Phi(v, L) \approx \sum_{\text{lattice: }M} w_M v(M^{-1}L)$. We
further approximate $v(M^{-1}L)$ with look-up at a lattice point
location.

Consider the case of rotation and translation invariance. More
intuitively, we stack rotated versions of the input images onto each
other in a $3$ dimensional space -- $2$ dimensions for the location of
a sample and $1$ dimension for the rotation of the image. A grid-based
convolution would not work here because the rotated image points might
not coincide with a grid anymore. Filtering in the permutohedral space
naturally respects the augmented feature space.

%% It is precisely for those somewhat low dimensional spaces where the
%% lattice is powerful. For higher dimensional space $k-d$ trees would be
%% better.

\section{Experiments}
\begin{table}
  \begin{center}
    \hspace{-3em}\begin{subtable}{.6\linewidth}
      \footnotesize
      \centering
      \vspace{0pt}
      \begin{tabular}[t]{lllll}
      \toprule
              &     & \multicolumn{3}{c}{Test Subsampling} \\
      Method  & Original & 100\% & 60\% & 20\%\\
      \midrule
      LeNet &  \textbf{0.9919} & 0.9660 & 0.9348 & \textbf{0.6434} \\
      PCNN &  0.9903 & \textbf{0.9844} & \textbf{0.9534} & 0.5767 \\
      \hline
      LeNet 100\% & 0.9856 & 0.9809 & 0.9678 & \textbf{0.7386} \\
      PCNN 100\% & \textbf{0.9900} & \textbf{0.9863} & \textbf{0.9699} & 0.6910 \\
      \hline
      LeNet 60\% & 0.9848 & 0.9821 & 0.9740 & 0.8151 \\
      PCNN 60\% & \textbf{0.9885} & \textbf{0.9864} & \textbf{0.9771} & \textbf{0.8214}\\
      \hline
      LeNet 20\% & \textbf{0.9763} & \textbf{0.9754} & 0.9695 & 0.8928 \\
      PCNN 20\% & 0.9728 & 0.9735 & \textbf{0.9701} & \textbf{0.9042}\\
      \bottomrule
      \end{tabular}
      \subcaption{Random Sampling}
      \label{tab:all-results}
    \end{subtable}
    \begin{subtable}{.28\linewidth}
      \centering
      \vspace{0pt}
      \footnotesize
      \begin{tabular}[t]{lr}
        \toprule
        Method & PSNR \\
        \midrule
        Noisy Input & $20.17$ \\
        CNN & $26.27$ \\
        PCNN Gauss & $26.51$ \\
        PCNN Trained & $26.58$ \\
        PCNN + CNN Trained & $26.65$ \\
        \bottomrule
      \end{tabular}
      \subcaption{Denoising}
      \label{tab:denoising}
    \end{subtable}
    % \vspace{-.3cm}
  \end{center}
  \caption{{\bfseries (a)} Classification accuracy on MNIST. We compare
    the LeNet~\citep{lecun98ieee} implementation that is part of
    Caffe~\citep{jia2014caffe} to the network with the first layer replaced by a
    permutohedral convolution layer (PCNN). Both are trained on the original image
    resolution (first two rows). Three more PCNN and CNN models are trained with
    randomly subsampled images (100\%, 60\% and 20\% of the pixels). An additional
    bilinear interpolation layer samples the input signal on a spatial grid for the
    CNN model. {\bfseries (b)} PSNR results of a denoising task using the BSDS500
    dataset~\citep{arbelaezi2011bsds500}.
  }
% \vspace{-.3cm}
\end{table}

% TODO add description for LeNet.

% \begin{table}
%   \begin{center}
%     %% \footnotesize
%     \centering
%     \begin{tabular}[t]{lllll}
%       LeNet & Permutohedral % & Encoded Rotation
%       & Subsample (100\%) & Subsample (60\%) & Subsample (20\%)\\
%       $0.9919$ & $0.9903$ & $0.9745$ & $0.9537$ & $0.8689$
%     \end{tabular}
%   \end{center}
%   \caption{Classification accuracy of the conducted experiments
%     compared to the LeNet~\citep{lecun98ieee} implementation that is
%     part of the caffe implementation~\citep{jia2014caffe}.}
%   \label{tab:all-results}
% \end{table}

We investigate the performance and flexibility of the proposed method
on two sets of experiments. The first setup compares the permutohedral
convolution with a spatial convolution that has been combined with a
bilinear interpolation. The second part adds a denoising experiment
to show the modelling strength of the permutohedral convolution.

It is natural to ask, how a spatial convolution combined with an
interpolation compares to a permutohedral convolutional neural network
(PCNN). The proposed convolution is particularly advantageous in cases
where samples are addressed in a higher dimensional space. %
% compared to only two dimensions in the spatial convolutional case.
Nevertheless, a bilinear interpolation prior to a spatial convolution
can be used for dense 2-dimensional positional features.
%The spatial convolution can operate on the regular grid
%structure that is create by the interpolation.

We take a reference implementation of LeNet~\citep{lecun98ieee} that
is part of the caffe project~\citep{jia2014caffe} on the MNIST dataset
as a starting point for the following experiments. The permutohedral
convolutional layer is also implemented in this framework.

We first compare the LeNet in terms of test-time accuracy when
substituting only the first of the convolutional layers with a
(position only) permutohedral layer and leave the rest identical.
\Cref{tab:all-results} shows that a similar performance is achieved,
so it seems model flexibility is not lost. The network is trained
according to the training parameters from the reference
implementation. % (0.9919 vs. 0.9903).
Next, we randomly sample continuous points in the input image, use
their interpolated values as signal and continuous positions as
features. Interestingly, we can train models with a different amount
of sub-sampling than at test time. The permutohedral representation is
robust with respect to this sparse input
signal. \Cref{tab:all-results} shows experiments with different signal
degradation levels. All the sampling strategies have in common that
the original input space of 28 by 28 pixels is densely covered. Hence,
a bilinear interpolation prior to the first convolution allows us to
compare against the original LeNet architecture. This baseline model
performs similar to a PCNN.

One of the strengths of the proposed method is that it does not depend
on a regular grid sampling as the tranditional convolution operators.
We highlight this feature with the following denoising experiment and change the
sampling space to be both sparse and 3-dimensional. The higher dimensional space
renders a bilinear interpolation and spatial convolution more and more
in-feasible due to the high number of corners of the hyper-cubical tessellation
of the space. We compare the proposed permutohedral convolution in an
illustrative denoising experiment to a spatial convolution. For bilateral
filtering, which is one of the algorithmic use-cases of the permutohedral
lattice, the input space features contain both the coordinates of a data sample
and the color information of the image; hence a 5-dimensional vector for color
images and a 3-dimensional vector for gray-scale images. In contrast to a direct
application of a bilateral convolution to the noisy input the filter for a
bilateral layer of a PCNN can now be trained. All experiments compare the
performance of a PCNN to a common CNN with images from the BSDS500
dataset~\citep{arbelaezi2011bsds500}. Each image is transformed into gray-scale
by taking the mean across channels and noise is artificially added to it with
samples from a Gaussian distribution $\mathcal{N}(\mu, \sigma^2), \mu = 0,
\sigma = \frac {25} {255}$.

The baseline network uses a spatial convolution (``CNN''
in~\Cref{tab:denoising}) with a kernel size of $5$ and predicts the scalar
gray-scale value at each pixel ($25$ filter weights). The layer is trained with
a fixed learning rate of $10^{-3}$, momentum weight of $0.9$ and a weight decay
of $5\cdot 10^{-4}$ on the ``train'' set. In the second architecture the
convolution is performed on the permutohedral lattice (``PCNN Gauss'' and ``PCNN
Trained'' in~\Cref{tab:denoising}). We include the pixel's gray value as an
additional feature for the generalized operation and set the neighborhood size
to $2$ ($65$ filter weights). The filter weights are initialized with a Gaussian
blur and are either applied directly to the noisy input (``PCNN Gauss'') or
trained on the ``train'' set to minimize the Euclidean distance to the clean
image with a learning rate of $0.1$. We cross-validate the scaling of the input
space on the ``val'' image set and reuse this setting for all experiments that
operate on the permutohedral lattice. A third architecture that combines both
spatial and permutohedral convolutions by summation (``CNN + PCNN'') is
similarly trained and tested.

We evaluate the PSNR utility averaged over the images from the ``test'' set and
see a slightly better performance of the bilateral network (``PCNN trained'')
with trained filters in comparison to a bilateral filter (``PCNN Gauss'') and
linear filter (``CNN''), see~\Cref{tab:denoising}. Both convolutional operations
combined further improve the performance and suggest that they have
complementary strengths. Admittedly this setup is rather simple, but it
validates that the generalized filtering has an advantage.

%% Our last experiment directly models domain knowledge of MNIST with the
%% new layer inside the network architecture. Without changing the data
%% itself we encode a rotation invariance of roughly 10 degrees. Training
%% the model with this kind of model yields a comparable performance of
%% $0.9893$ (versus $0.9919$ LeNet).  We conducted these experiments as
%% proof of concept, in the future we will investigate other invariances
%% for example scale invariance.

In the future we plan to investigate the use of the PCNN architecture
for other computer vision problems, e.g. semantic segmentation, and
modeling domain knowledge like rotation or scale invariance.

\section{Conclusion}

This paper presents a generalization of the convolutional operation to
sparse input signals. We envision many consequences of this work.
Consider signals that are naturally represented as measurements
instead of images, like MRT scan readings. The permutohedral lattice
filtering avoids the pre-processing assembling operation into a dense
image, it is possible to work on the measured sparse signal directly.
Another promising use of this filter is to encode scale invariance,
typically this is encoded by presenting multiple scaled versions of an
image to several branches of a network. The convolution presented here
can be defined on the continuous range of image scales without a
finite subselection. In summary, this technique allows to encode prior
knowledge about the observed signal to define the domain of the
convolution. The typical spatial filter of CNNs is a particular type
of prior knowledge, we generalize this to sparse signals.

\bibliography{iclr2015}

\begin{thebibliography}{10}
\providecommand{\natexlab}[1]{#1}
\providecommand{\url}[1]{\texttt{#1}}
\expandafter\ifx\csname urlstyle\endcsname\relax
  \providecommand{\doi}[1]{doi: #1}\else
  \providecommand{\doi}{doi: \begingroup \urlstyle{rm}\Url}\fi

\bibitem[Adams et~al.(2009)Adams, Gelfand, Dolson, and Levoy]{adams09siggraph}
Adams, Andrew, Gelfand, Natasha, Dolson, Jennifer, and Levoy, Marc.
\newblock {G}aussian kd-trees for fast high-dimensional filtering.
\newblock In \emph{ACM SIGGRAPH 2009 Papers}, SIGGRAPH '09, pp.\  21:1--21:12,
  New York, NY, USA, 2009.

\bibitem[Adams et~al.(2010)Adams, Baek, and Davis]{adams10eurographics}
Adams, Andrew, Baek, Jongmin, and Davis, Myers~Abraham.
\newblock Fast high-dimensional filtering using the permutohedral lattice.
\newblock \emph{Comput. Graph. Forum}, 29\penalty0 (2):\penalty0 753--762,
  2010.

\bibitem[Arbeláez et~al.(2011)Arbeláez, Maire, Fowlkes, and
  Malik]{arbelaezi2011bsds500}
Arbeláez, Pablo, Maire, Michael, Fowlkes, Charless, and Malik, Jitendra.
\newblock Contour detection and hierarchical image segmentation.
\newblock \emph{IEEE Trans. Pattern Anal. Mach. Intell.}, 33\penalty0
  (5):\penalty0 898--916, May 2011.

\bibitem[Aurich \& Weule(1995)Aurich and Weule]{aurich95dagm}
Aurich, Volker and Weule, J{\"{o}}rg.
\newblock Non-linear {G}aussian filters performing edge preserving diffusion.
\newblock In \emph{Mustererkennung 1995, 17. DAGM-Symposium, Bielefeld, 13.-15.
  September 1995, Proceedings}, pp.\  538--545, 1995.

\bibitem[Jia et~al.(2014)Jia, Shelhamer, Donahue, Karayev, Long, Girshick,
  Guadarrama, and Darrell]{jia2014caffe}
Jia, Yangqing, Shelhamer, Evan, Donahue, Jeff, Karayev, Sergey, Long, Jonathan,
  Girshick, Ross, Guadarrama, Sergio, and Darrell, Trevor.
\newblock Caffe: Convolutional architecture for fast feature embedding.
\newblock \emph{arXiv preprint arXiv:1408.5093}, 2014.

\bibitem[LeCun et~al.(1998)LeCun, Bottou, Bengio, and Haffner]{lecun98ieee}
LeCun, Yann, Bottou, L{\'{e}}on, Bengio, Yoshua, and Haffner, Patrick.
\newblock Gradient-based learning applied to document recognition.
\newblock \emph{Proceedings of the IEEE}, 86\penalty0 (11):\penalty0
  2278--2324, November 1998.

\bibitem[Mallat(2012)]{mallat12pam}
Mallat, St{\'{e}}phane.
\newblock Group invariant scattering.
\newblock \emph{Communications in Pure and Applied Mathematics}, 65\penalty0
  (10):\penalty0 1331--1398, 2012.

\bibitem[Paris \& Durand(2009)Paris and Durand]{paris09ijcv}
Paris, Sylvain and Durand, Fr{\'e}do.
\newblock A fast approximation of the bilateral filter using a signal
  processing approach.
\newblock \emph{International Journal of Compututer Vision}, 81\penalty0
  (1):\penalty0 24--52, January 2009.

\bibitem[Smith \& Brady(1997)Smith and Brady]{smith97ijcv}
Smith, Stephen~M. and Brady, J.~Michael.
\newblock {SUSAN} -- a new approach to low level image processing.
\newblock \emph{Int. J. Comput. Vision}, 23\penalty0 (1):\penalty0 45--78, May
  1997.
\newblock ISSN 0920-5691.

\bibitem[Tomasi \& Roberto(1998)Tomasi and Roberto]{tomasi98iccv}
Tomasi, Carlo and Roberto, Manduchi.
\newblock Bilateral filtering for gray and color images.
\newblock In \emph{Proceedings of the Sixth International Conference on
  Computer Vision}, ICCV '98, pp.\  839--846, Washington, DC, USA, 1998. IEEE
  Computer Society.

\end{thebibliography}
\bibliographystyle{iclr2015}

\end{document}